%% file: ijcai21-multiauthor.tex
\documentclass{article}
\usepackage{ijcai21}

\usepackage{times}

\usepackage{soul}
\usepackage{url}
\usepackage[hidelinks]{hyperref}
\usepackage[utf8]{inputenc}
\usepackage[small]{caption}
\usepackage{graphicx}
\usepackage{amsmath}
\usepackage{booktabs}
\input{preamble/math_commands.tex}

\input{preamble/new_commands.tex}

\usepackage{url}
\usepackage{natbib}
\usepackage[T1]{fontenc}    %
\usepackage{amsfonts}       %
\usepackage{nicefrac}       %
\usepackage{microtype}      %
\usepackage{url}
\usepackage[noend]{algpseudocode}
\usepackage{float}
\usepackage{wrapfig,lipsum,booktabs}
\usepackage{enumitem}
\usepackage[font=small,skip=5pt]{caption}
\usepackage{subcaption}

\newsavebox{\measurebox}

\usepackage[colorinlistoftodos]{todonotes}
\usepackage{comment}

\usepackage{array,ragged2e}
\newcolumntype{P}[1]{>{\RaggedRight\arraybackslash}p{#1}}

\newcommand{\Tableref}[1]{Table \ref{#1}}

\usepackage[compact]{titlesec} %

\definecolor{revision_color}{HTML}{397517}
\definecolor{revision_color2}{HTML}{294FBB}

\renewenvironment{description}[1][0pt]
  {\list{}{\labelwidth=0pt \leftmargin=#1
   }}
  {\endlist}

\title{Reinforcement Learning for Sparse-Reward Object-Interaction Tasks in a First-person Simulated 3D Environment}

\author{
Wilka Carvalho$^1$\footnote{
  Contact author: \texttt{wcarvalh@umich.edu}}\and
Anthony Liang$^1$\and
Kimin Lee$^{2}$\and
Sungryull Sohn$^1$\and \\
Honglak Lee$^1$\and
Richard Lewis$^1$\And
Satinder Singh$^1$
\\
\affiliations
$^1$University of Michigan \\
$^2$UC Berkeley
}

\begin{document}

\maketitle
\begin{abstract}

\input{main/abstract.tex}

\end{abstract}

  \section{Introduction}
  \input{main/introduction.tex}

  \section{Related Work}
  \input{main/related_work.tex}

  \section{Sparse-Reward Object-Interaction Tasks in a First-Person Simulated 3D environment}
  \input{main/preliminaries.tex}

  \section{LOAD: Learning Object Attention \& Dynamics Agent}
  \input{main/methods.tex}

  \section{Experiments}

\input{main/experiments.tex}

  \section{Conclusion}
  \input{main/conclusion.tex}

  \bibliographystyle{named}
  \bibliography{bib.bib}

  \appendix
  \clearpage
  \newpage

  \section{Agent Details} \label{appendix:implementation}

  \subsection{Architectures and objective functions}
  \input{supplementary/architectures.tex} \label{appendix:architecture}
  
  \subsection{Hyperparameter Search} \label{appendix:hyperparameters}
  \input{supplementary/hyperparameter.tex}

  \section{Thor Implementation Details} \label{appendix:task-details} \label{appendix:thor-settings}
  \subsection{Thor Settings}
  \input{supplementary/thor-settings.tex}
  \subsection{Task Details}
  \input{supplementary/task_details.tex}

  \subsection{Interaction Dataset}

\input{supplementary/datasets.tex}

  \section{Additional Results}\label{sec:additional}
  \input{supplementary/additional_results.tex}

\end{document}

%% file: preamble/math_commands.tex
\usepackage{amsmath,amsfonts,bm}

\def\Figref#1{Figure~\ref{#1}}

\def\eqref#1{equation~\ref{#1}}

\def\1{\bm{1}}

\def\vz{{\bm{z}}}

\def\mZ{{\bm{Z}}}

\DeclareMathAlphabet{\mathsfit}{\encodingdefault}{\sfdefault}{m}{sl}
\SetMathAlphabet{\mathsfit}{bold}{\encodingdefault}{\sfdefault}{bx}{n}

\newcommand{\E}{\mathbb{E}}

\newcommand{\R}{\mathbb{R}}

\DeclareMathOperator*{\argmax}{arg\,max}
\DeclareMathOperator*{\argmin}{arg\,min}

%% file: preamble/new_commands.tex
\newcommand{\reviewer}[3]{
	\expandafter\newcommand\csname #1\endcsname[1]{
		\textcolor{#3}{[#2: ##1]}
	}
}

\reviewer{kimin}{Kimin}{blue}

\newcommand{\twoline}[2]{\begin{tabular}{@{}c@{}}#1 \\ #2\end{tabular}}

\newcommand{\attention}{A}

%% file: main/abstract.tex
Learning how to execute complex tasks involving multiple objects in a 3D world is challenging when there is no ground-truth information about the objects or any demonstration to learn from.
When an agent only receives a signal from task-completion, this makes it challenging to learn the object-representations which support learning the correct object-interactions needed to complete the task.
In this work, we formulate learning an attentive object dynamics model as a classification problem, using random object-images to define incorrect labels. 
We show empirically that this enables object-representation learning that captures an object's category (is it a toaster?), its properties (is it on?), and object-relations (is something inside of it?).
With this, our core learner (a relational RL agent) receives the dense training signal it needs to rapidly learn object-interaction tasks.
We demonstrate results in the 3D AI2Thor simulated kitchen environment with a range of challenging food preparation tasks.
We compare our method's performance to several related approaches and against the performance of an oracle: an agent that is supplied with ground-truth information about objects in the scene.
We find that our agent achieves performance closest to the oracle in terms of both learning speed and maximum success rate.

%% file: main/introduction.tex
Consider a robotic home-aid agent that learns \emph{object-interaction tasks} that involve using multiple objects together to accomplish various tasks such as chopping vegetables or heating meals. 
Such tasks are important for artificial intelligence (AI) to make progress on because of their large potential to impact our everday world: 
  nursing robots can serve healthcare workers in hospitals,
  and home-aid robots can help busy families, the disabled, and the elderly.

Prior work on object-interaction tasks has focused on achieving strong training performance using expert demonstrations~\citep{zhu2017visual, Shridhar2019ALFREDAB}. Unfortunately,~\citet{zhu2017visual} found they were unable to learn relatively simple pick and place tasks when only learning from a sparse task-completion signal. Other work has relaxed the learning problem by relying on domain knowledge in the form of shaped rewards or object-affordance knowledge~\citep{jain2019two, gordon2018iqa}. 

Unfortunately, expert demonstrations and shaped rewards can be challenging to obtain for tasks novel to an agent. Additionally, it can be tedious or impossible to obtain ground-truth information about all novel objects an agent may encounter. Ideally, agents are capable of learning object-interaction tasks without this information. To work towards this, we focus on the setting where none of these are available.

Learning object-interaction tasks without expert demonstrations or shaped rewards is challenging because selecting between object-interactions induces a branching factor that scales with the number of visible objects, leading the agent choose from 50-100 actions at a given time-step. This leads the agent to infrequently experience a successful episode. When the agent does, task completion typically occurs after many hundred time-steps. Consider learning to toast bread. The agent should learn to turn on the toaster after a bread slice is placed inside, i.e. it needs to learn to represent \emph{containment relationships} (the bread is inside the toaster) and \emph{object properties} (the toaster is on or off). Without domain knowledge about objects, task-completion alone provides a weak learning signal for learning both to represent 3D object categories, properties, and relationships. When episodes last for hundreds of time-steps and the agent interacts with many objects, this makes it challenging to learn about about how the agent's object-interactions led to reward.

In this work, we find that we can achieve strong training performance on object-interaction tasks without expert demonstrations, shaped rewards, or ground-truth object-knowledge by incorporating inter-object attention and an object-centric model into a reinforcement learning agent.We call our agent the \emph{Learning Object Attention \& Dynamics} (or \emph{LOAD}) agent. 
LOAD is composed of a base object-centric relational policy (\emph{Attentive Object-DQN}, \S \ref{sec:relational-dqn}) that leverages inter-object attention to incorporate object-relationships when estimating object-interaction action-values.
Without ground-truth information to identify object categories, properties, or relationships, LOAD learns object-representations with a novel learning objective that frames learning an object-model as a classification problem, where random object-embeddings are incorrect labels (\emph{Attentive Object-Model}, \S \ref{sec:relational-model}). By doing so, we provide the object-model with a dense learning signal for learning represent both object categories, but also changes in object-properties caused by different object-interactions.
Additionally, by sharing inter-object attention between the policy and the model, learning the model helps drive learning of inter-object attention helpful for speedening task learning. 

In order to study object-interaction tasks and evaluate our agent, we adopt the virtual home-environment AI2Thor \citep{kolve2017ai2thor} (or \emph{Thor}). Thor is an open-source environment that is high-fidelity, 3D, partially observable, and enables object-interactions. We show that LOAD is able to significantly reduce sample complexity in this domain where no prior work has yet learned sparse-reward object-interaction tasks without expert demonstrations or shaped rewards.

In our main evaluation, we compare pairing Attentive Object-DQN with our Attentive Object-Model to alternative representation learning methods, and show that learning with our object-model best closes the performance gap to an agent supplied with ground-truth information about object categories, properties, and relationships (\S \ref{sec:results}). Through an analysis of the learned object-representations and inter-object attention learned by each auxiliary task, we provide quantitative evidence that our Attentive Object-Model best learns representations that capture the ground-truth information present in our oracle (\S \ref{sec:rep-learning}). We hypothesize that this is the source of our strong performance. Afterwards, we perform a series of ablations to study the importance of object-representations which capture object-properites and object-relations for reducing sample-complexity (\S \ref{sec:attention-ablation}).

In summary, the key contributions of our proposal are: (1) LOAD: an RL agent that demonstrates how to learn sparse-reward object-interaction tasks with first-persosn vision without expert demonstrations, shaped rewards, or ground-truth object-knowledge. (2) A novel Attentive Object-Model auxilliary task, which frames learning an object-model as a classification problem. With our analysis, we provide evidence that for our 3D, high-fidelty domain and our architecture, it is key to learn object-representations which not only capture object-categories but also object-properties and object-relations.

%% file: main/related_work.tex
\textbf{Learning object-interaction tasks in 3D, first-person environments}.
Due to the large branching factor induced by object-interactions, most work here has relied extensively on expert demonstrations~\citep{zhu2017visual, Shridhar2019ALFREDAB,xu2019regression} or avoided this problem by hard-coding object-selection~\citep{jain2019two,gordon2018iqa}.
The work most closely related to ours is \citet{oh2017zero} (in Minecraft) and \citet{zhu2017visual} (in Thor). 
Both develop a hierarchical reinforcement learning agent where a meta-controller provides goal object-interactions for a low-level controller to complete using ground-truth object-information. 
Both provide agents with knowledge of all objects and both assume lower-level policies pretrained to navigate to objects and to select interactions with a desired object.  In contrast, we do not provide the agent with any ground-truth object information; nor do we pretrain navigation to objects or selection of them.

\textbf{Object-Centric Relational RL}.
An intutive approach to tasks with objects is object-centric relational RL. Most work here has used hand-designed representations of objects and their relations, showing things like improved sample-efficiency~\citep{xu2020learning}, improved policy quality~\citep{zaragoza2010relational}, and generalization to unseen objects~\citep{van2015learning}. In contrast, we seek to learn object-representations and object-relations implicitly with our network. 
Most similar to our work is \citet{zambaldi2018relational}--which applies attention to the feature vector outputs of a CNN. 
In this work, Attentive Object-DQN is a novel architecture extension for a setting with an object-centric observation- and action-space. Additionally, we show that learning an object-model as an auxilliary task can help drive learning of attention.

\textbf{Learning an object-model as an auxiliary task}.
Most prior work here has focused on how an object-model can be used in model-based reinforcement learning by enabling superior planning~\citep{ye2020object, veerapaneni2020entity, watters2019cobra}. 
In contrast, we do not use our object-model for planning and instead show that it can be leveraged to learn object-representation and inter-object attention to support faster policy learning in a model-free setting. Additionally, other work focused on domains where representation-learning only had to differntiate object-categories. We show that our method can additionally differentiate object-properties and does so significantly better than the object-model of ~\citet{watters2019cobra}.
Our attentive object-model is most similar to the Contrastive Structured World Model (CSWM)~\citep{kipf2019contrastive}, which uses a maximum margin contrastive learning objective~\citep{hadsell2006dimensionality} to learn an object-model. Instead, we formulate a novel object-model contrastive objective as learning a classification problem. 
We note that they applied their model towards video-prediction and not reinforcement learning.

%% file: main/preliminaries.tex
\begin{figure}[!htb]
  \centering
  \includegraphics[width=0.45\textwidth]{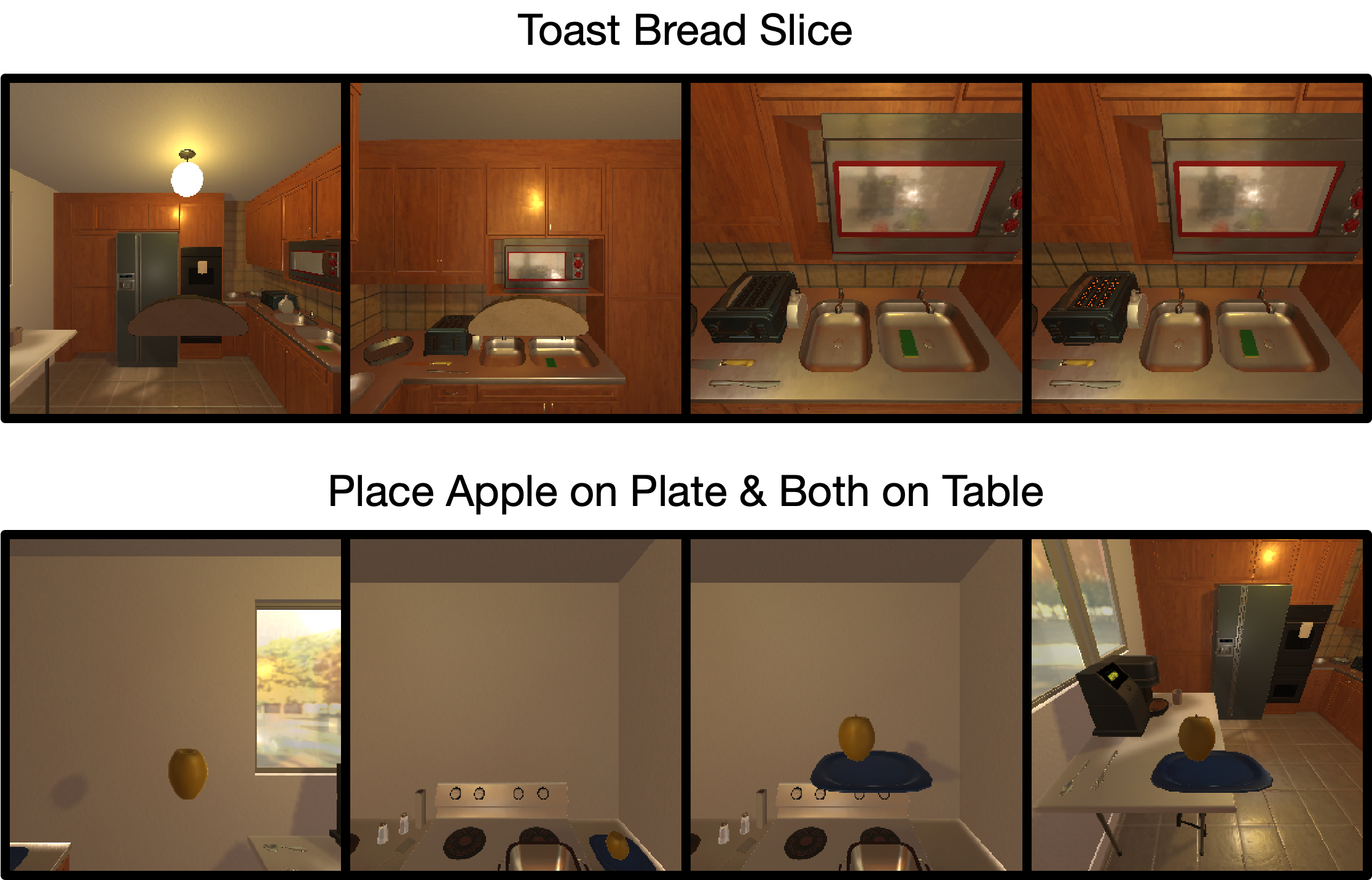}
  \caption{We present the steps required to complete two of our tasks. 
  In ``Toast Bread Slice'', an agent must pickup a bread slice, bring it to the toaster, place it in the toaster, and turn the toaster on. In order to complete the task, the agent needs to recognize the toaster across angles, and it needs to recognize that when the bread is inside the toaster, turning the toaster on will cook the bread. 
  In ``Place Apple on Plate \& Both on Table'', agent must pickup an apple, place it on a plate, and move the plate to a table. It must recognize that because the objects are combined, moving the plate to the table will also move the apple. 
  We observe that learning to use objects together such as in the tasks above poses a representation learning challenge -- and thus policy learning challenge -- when learning from only a task-completion reward. }
  \label{fig:task-completion}
\end{figure}

\begin{table*}[!h]
  \input{main/tasks.tex}
  \caption{Description of challenges associated with the tasks we study. See \Figref{fig:task-completion} for example panels of 2 tasks.}
  \label{table:task-challenges}
\end{table*}

\label{sec:framework}
\textbf{Observations}. We focus on an agent that has a 2D camera for experiencing \textit{egocentric} observations $x^{\tt ego}$ of the environment. Our agent also has a pretrained vision system that enables it to extract bounding box image-patches corresponding to the visible objects in its observation $X^o=\{x^{o,i}\}$. Besides boxes around objects, no other information is extracted (i.e. no labels, identifiers, poses, etc.). We assume the agent has access to its $(x,y,z)$ location and body rotation $(\varphi_1, \varphi_2, \varphi_3)$ in a global coordinate frame, $x^{\tt loc}=(x,y,z,\varphi_1, \varphi_2, \varphi_3)$. 

\textbf{Actions}.
In this work, we focus on the Thor environment. Here, the agent has $8$ base object-interactions: $\mathcal{I}=$ \{\textit{Pickup, Put, Open, Close, Turn on, Turn off, Slice, Fill}\}. 
The agent interacts with objects by selecting (object-image-patch, interaction) pairs $a=(b, x^{o,c}) \in \mathcal{I} \times X^o$, where $x^{o,c}$ corresponds to the \textit{chosen} image-patch.
For example, the agent can turn on the stove by selecting the image-patch containing the stove-knob and the \emph{Turn on} interaction (see Figure \ref{fig:full-architecture} for a diagram). 
Each action is available at every time-step and can be applied to all objects (i.e. no affordance information is given/used).
Interactions occur over one time-step, though their effect may occur over multiple.
For the example above, when the agent applies ``Turn on'' to the stove knob, food on the stove will take several time-steps to heat.

In addition to object-interactions, the agent can select from 8 base navigation actions: $\mathcal{A}_N=$ \{\textit{Move ahead, Move back, Move right, Move left, Look up, Look down, Rotate right, Rotate left}\}. With \{\textit{Look up, Look down}\}, the agent can rotate its head up or down in increments of $30^{\circ}$ between angles $\{0^{\circ}, \pm 30^{\circ}, \pm  60^{\circ}\}$. $0^{\circ}$ represents looking straight ahead. With \{\textit{Rotate Left, Rotate Right}\}, the agent can rotate its body by $\{\pm 90^o\}$.

\textbf{Tasks}.
We construct 8 tasks with the following 4 challenges.
Challenge (A): the visual complexity of task objects (e.g. the cup is translucent).
Challenge (B): the number of objects to be interacted with (e.g., ``Slice Apple, Potato, Lettuce'' requires the agent interact with 4 objects).
Challenge (C): whether object-containment must be recognized and used (e.g. toasting bread in a toaster). 
Challenge (D): whether object-properties change (e.g. bread get's cooked). 
See \Figref{table:task-challenges} for a description of the challenges associated with each task and \Figref{fig:task-completion} for example panels of 2 tasks. 

\textbf{Reward}. We consider a single-task setting where the agent receives a terminal reward of $1$ upon task-completion.

%% file: main/tasks.tex
  \resizebox{\textwidth}{!}{%
  \begin{tabular}{|P{3.25cm}|P{3.25cm}|P{3.25cm}|P{3.25cm}|P{3.25cm}|P{3.25cm}|}
    \toprule
    Slice $\{X_i\}$, $n \in [1,3]$ & Make Toamto \& Lettuce Salad & Place Apple on Plate, Both on Table & Cook Potato on Stove & Fill Cup with Water & Toast Bread Slice   \\
    \midrule

      (A) recognize knife across angles\newline
      (B) recognize 2-4 objects\newline
    
    & 

      (B) recognize 3 objects \newline
      (C) use containment: plate with tomato/lettuce slice

    & 

      (B) recognize 3 objects \newline
      (C) use containment: apple on plate

    & 

      (B) recognize 2 objects \newline
      (C) use containment: potato on stove \newline
      (D) changing properites: cooked potato

    & 

      (A) recognize translucent cup across backgrounds \newline
      (B) recognize 2 objects \newline
      (C) use containment: cup in sink \newline
      (D) changing properites: filled cup

    & 

      (A) recognize toaster across angles \newline
      (B) recognize 2 objects \newline
      (C) use containment: bread inside toaster \newline
      (D) changing properites: cooked bread

    \\
    \midrule
  \end{tabular}
  }

%% file: main/methods.tex
LOAD is a reinforcement learning agent composed of an object-centric relational policy, Attentive Object-DQN, and an Attentive Object-Model.
LOAD uses 2 perceptual modules. The first, $f_{\tt enc}^o$, takes in an observation $x$ and produces object-encodings $\{\vz^{o,i}\}_{i=1}^n$ for the $n$ visible object-image-patches $X^o = \{x^{o,i}\}_{i=1}^n$, where $\vz^{o,i} \in \R^d_o$. The second, $f_{\tt enc}^{\kappa}$, takes in the egocentric observation and location $x^{\kappa}=[x^{\tt ego}, x^{\tt loc}]$ to produce the \textit{context} for the objects $\vz^{\kappa} \in \R^d_{\kappa}$.
LOAD treats state as the union of these variables: $s=\{\vz^{o,i}\}\cup \{\vz^{\kappa}\}$.
Given object encodings, Attentive Object-DQN computes action-values $Q(s, a=(b, x^{o,i}))$ for interacting with an object $x^{o,i}$ and leverages an attention module $\attention$ to incorporate information about other objects $x^{o,j \neq i}$ into this computation (see \S \ref{sec:relational-dqn}).

\begin{figure*}[!t]
  \centering
  \includegraphics[width=1\textwidth]{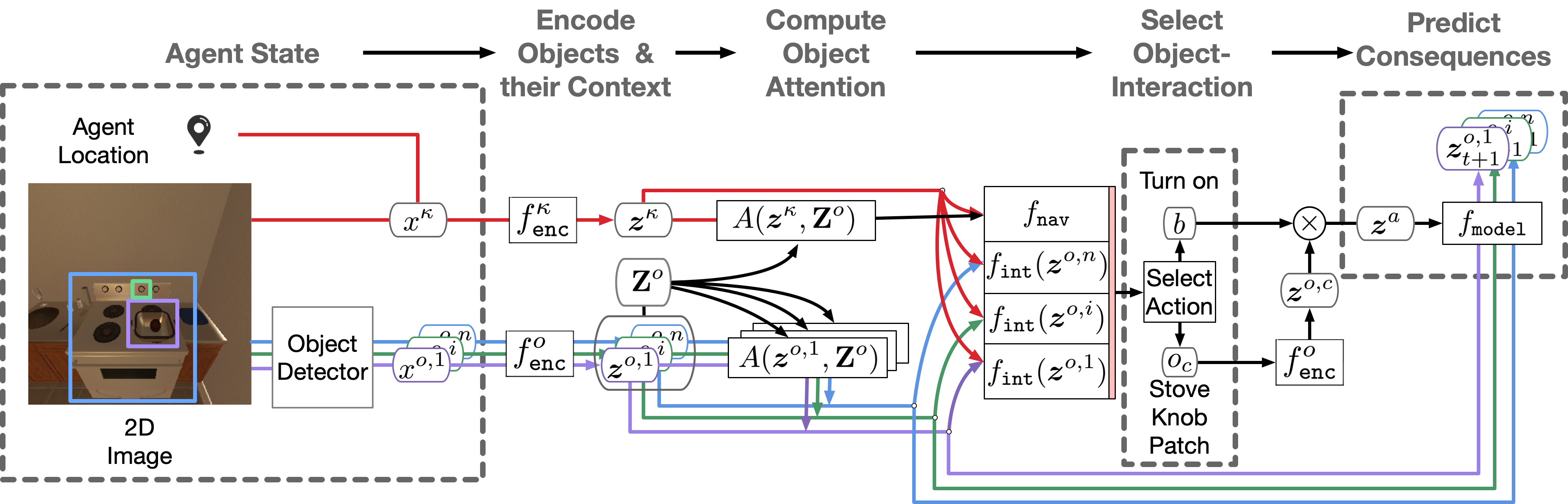}
  \caption{Full architecture and processing pipeline of LOAD. A scene is broken down into object-image-patches $\{x^{o,j}\}$ (e.g. of a pot, potato, and stove knob). The scene image is combined with the agent's location to define the \textit{context} of the objects, $x^\kappa$. The objects  $\{x^{o,j}\}$ and their context $x^\kappa$ are processed by different encoding branches and then recombined by an attention module $\attention$ that selects relevant objects for computing Q-value estimates. Here, $\attention$ might select the pot image-patch when computing $Q$-values for interacting with the stove-knob image-patch. Actions are selected as (object-image-patch, base action) pairs $a=(b, x^{o,c})$. The agent then predicts the consequences of its interactions with our attentive object-model $f_{\tt model}$ which reuses $\attention$.}
  \label{fig:full-architecture}
\end{figure*}

To address the representation learning challenge induced by a sparse-reward signal, object-representations $\vz^{o,i}$ and object-attention $\attention$ are trained to predict object-dynamics with an attentive object-model (see \S \ref{sec:relational-model}).
See Figure~\ref{fig:full-architecture} for an overview of the full architecture.

\subsection{Attentive Object-DQN} 
\label{sec:relational-dqn}
\input{main/relational-object-dqn.tex}

\subsection{Attentive Object-Dynamics Model}
\label{sec:relational-model}
\input{main/attentive-object-model.tex}

%% file: main/relational-object-dqn.tex
Attentive Object-DQN uses $\smash{\widehat{Q}(s,a)}$ to estimate the action-value function $\smash{Q^{\pi}(s,a)=}\linebreak{\E_{\pi} [\sum^{\infty}_{t=0} \gamma^t r_{t}|S_t=s, A_t=a]}$, which maps state-action pairs to the expected return on starting from that state-action pair and following policy $\pi$ thereafter. 

\textbf{Leveraging inter-object attention during action-value estimation}. In many tasks, an agent must integrate information about multiple objects when estimating $Q$-values. 
For example, in the ``toast bread'' task, the agent needs to integrate information about the toaster and the bread when deciding to turn on the toaster. 
To accomplish this, we exploit the object-centric observations-space and employ attention~\citep{vaswani2017attention} to incorporate inter-object attention into $Q$-value estimation.

More formally, given an object-encoding $\vz^{o,i}$, we can use attention to select relevant objects $\attention(\vz^{o,i}, \mZ^o) \in \R^{d_o}$ for estimating $Q(s, a=(b, x^{o,i}))$. With a matrix of object-encodings, $\mZ^o = \left[ \vz^{o,i} \right]_i \in \R^{n \times d_o}$, we can perform this computation efficiently for each object-image-patch via:
\begin{align} \label{eq:attention}
  \left(\begin{array}{c} \attention(\vz^{o,1}, \mZ^o) \\ \vdots \\ \attention(\vz^{o,n}, \mZ^o) \end{array}\right) = 
  \operatorname{ Softmax } \left(\frac{\left(\mZ^o W^{q_o}\right) \left(\mZ^o W^k \right)^{\top}}{\sqrt{d_k}}\right)\mZ^o.
\end{align}
Here, $\mZ^o W^{q_o}$ projects each object-encoding to a ``query'' space and $\mZ^o W^k$ projects each encoding to a ``key'' space, where their dot-product determines whether a key is selected for a query. The softmax acts as a soft selection-mechanism for selecting an object-encoding in $\mZ^o$. 

\textbf{Estimating action-values}. We can incorporate attention to estimate $Q$-values for selecting an interaction $b \in \mathcal{I}$ on an object $x^{o,i}$ as follows:
\begin{align} \label{eq:q-function}
  \widehat{Q}(s, a=(b, x^{o,i})) = f_{\tt int}([\vz^{o,i}, \attention(\vz^{o,i}, \mZ^o), \vz^{\kappa}])
\end{align}
Importantly, this enables us to compute $Q$-values for a variable number of \textit{unlabeled} objects. We can similarly incorporate attention to compute $Q$-values for navigation actions by replacing $\mZ^o W^{q_o}$ with $\left(W^{q_\kappa}\vz^{\kappa}\right)^\top$ in \eqref{eq:attention}.
We estimate $Q$-values for navigation actions $b \in \mathcal{A}_{N}$ as follows:
\begin{align} \label{eq:q-function-nav}
  \widehat{Q}(s, a=b) = f_{\tt nav}([\vz^{\kappa}, \attention(\vz^{\kappa}, \mZ^o)]) 
\end{align}

\textbf{Learning}. We estimate $\widehat{Q}(s,a)$ as a Deep Q-Network (DQN) by minimizing the following temporal difference objective:
\begin{equation} \label{eq:dqn_loss}
    \mathcal{L}_{\tt DQN} = \E_{s_t, a_t, r_t, s_{t+1}} \left[||y_t - \widehat{Q}(s_t,a_t; \theta) ||^2\right],
  \end{equation}
where $y_t = r_t + \gamma \widehat{Q}(s_{t+1},a_{t+1}; \theta_{\tt old})$ is the target Q-value, and $\theta_{\tt old}$ is an older copy of the parameters $\theta$. 
To do so, we store trajectories containing transitions $(s_t,a_t,r_t,s_{t+1})$ in a replay buffer that we sample from~\cite{mnih2015human}. 
To stabilize learning, we use Double-Q-learning~\cite{van2016deep} to choose the next action: $a_{t+1} = \argmax_a \widehat{Q}(s_{t+1}, a ; \theta)$.

%% file: main/attentive-object-model.tex
Consider the global set of objects $\smash{\{o^g_{t,i}\}^m_{i=1}}$, where $m$ is the number of objects in the environment.
At each time-step, each object-image-patch the agent observes corresponds to a 2D projection of $o^g_{t,i}$, $\smash {\rho(o^g_{t,i})}$ (or $\smash{\rho^{g,i}_{t}}$ for short) and encodes it as $\vz^{g,i}_{t}$. Given, an object-image-patch encoding $\smash{\vz^{g,i}_{t}}$ and a performed interaction $a_t$, we can define an object-dynamics model $\smash{D(\mZ^o_{t}, \vz^{g,i}_{t}, a_t)}$ which produces the resultant encoding for $\rho^{g,i}_{t+1}$.
We want $\smash{D(\mZ^o_{t}, \vz^{g,i}_{t}, a_t)}$ to be closer to $\vz^{g,i}_{t+1}$ than to encodings of other object-image-patches.

\textbf{Classification problem}. We can formalize this by setting up a classification problem.
For an object-image-patch encoding $\smash{\vz^{g,i}_{t}}$, we define the \emph{prediction} as the output of our object-dynamics model $\smash{D(\mZ^o_{t}, \vz^{g,i}_{t}, a_t)}$.
We define the \emph{label} as the encoding of a visible object-image-patch at the next time-step with the highest cosine similarity to the original encoding $\smash{\vz^{g,i}_{+} = \argmax_{z_{t+1}^{g,j}}}$ $ \cos(\vz_{t}^{g,i}, \vz_{t+1}^{g,j})$.
We can then select $K$ random object-encodings $\{\vz^{o}_{k, -}\}^K_{k=1}$ as \emph{incorrect labels}. Rewriting $D(\mZ^o_{t}, \vz^{g,i}_{t}, a_t)$ as $D$, this leads to:
\begin{align} \label{eq:forward-classification}
  p(\vz^{g,i}_{t+1}| \mZ^o_{t}, a_t) 
    &= \frac{\exp(D^{\top}\vz^{g,i}_{+})}
      {\exp(D^{\top}\vz^{g,i}_{+}) + \sum_k \exp(D^{\top}\vz^{o}_{k, -})}.
\end{align}
The set of indices corresponding to visible objects at time $t$ is $\smash{v_t = \{i: \rho^{g,i}_{t} \text{ is visible at time } t\}}$.
The set of observed object-image-patch encodings is then $\smash{ \mZ^o_t=\{\vz^{o,j}_{t}\}=\{\vz^{g,i}_{t}\}_{i \in v_t}}$.
Assuming the probability of each object's next state is conditionally independent given the current set of objects and the action taken, we arrive at the following objective:
\begin{align}
  \begin{split}\label{eq:forward-loss}
    \mathcal{L}_{\tt model} 
      &=  \E_{z_t, a_t, z_{t+1}} \left[ - \log p(\mZ^o_{t+1}| \mZ^o_{t}, a_t) \right] \\
      &= \E_{z_t, a_t, z_{t+1}} \left[-\sum_{i\in v_{t+1}} \log p(z^{g,i}_{t+1}| \mZ^o_{t}, a_t) \right]. \\
  \end{split}
\end{align}
Our final objective becomes:
\begin{equation}
  \mathcal{L} = \mathcal{L}_{\tt DQN}  + \beta^{\tt model} \mathcal{L}_{\tt model}.
\end{equation}

\textbf{Leveraging inter-object attention for improved accuracy.} 
Consider slicing an apple with a knife. When selecting ``slice'' on the apple patch, learning to attend to the knife patch \textbf{both} enables more accurate estimation of $Q$-values and higher model-prediction accuracy. 
We can accomplish this by incorporating $\attention(\vz^{g,i}, \mZ^o)$ into our object-model as follows:
\begin{align} \label{eq:relational-object-dynamics model}
  D(\mZ^o_{t}, \vz^{g,i}_{t}, a_t) &= f_{\tt model}([\vz_t^{g,i}, \attention(\vz_t^{g,i}, \mZ_t^o), \vz_t^a]).
\end{align}
To learn an action encoding $\vz_t^a$ for action $a_t$, following~\citet{oh2015action,reed2014learning}, we employ multiplicative interactions so our learned action representation $\vz_t^a$ compactly models the cartesian product of all base actions $b$ and object-image-patch selections $o_c$ as 
\begin{align} \label{eq:object-interaction}
  \vz_t^a &= W^{o}\vz_t^{g,c} \odot W^{b}b_t,
\end{align}
where $W^{o} \in \R^{d_a \times d_o}$, $W^{b} \in \R^{d_a \times |\mathcal{A}_I|}$, and $\odot$ is an element-wise hadamard product.
In practice, $f_{\tt model}$ is a small 1- or 2-layer neural network making this method compact and simple to implement.

%% file: main/experiments.tex
\begin{figure*}
  \begin{minipage}[b]{0.65\linewidth}
    \centering
    \includegraphics[width=\textwidth]{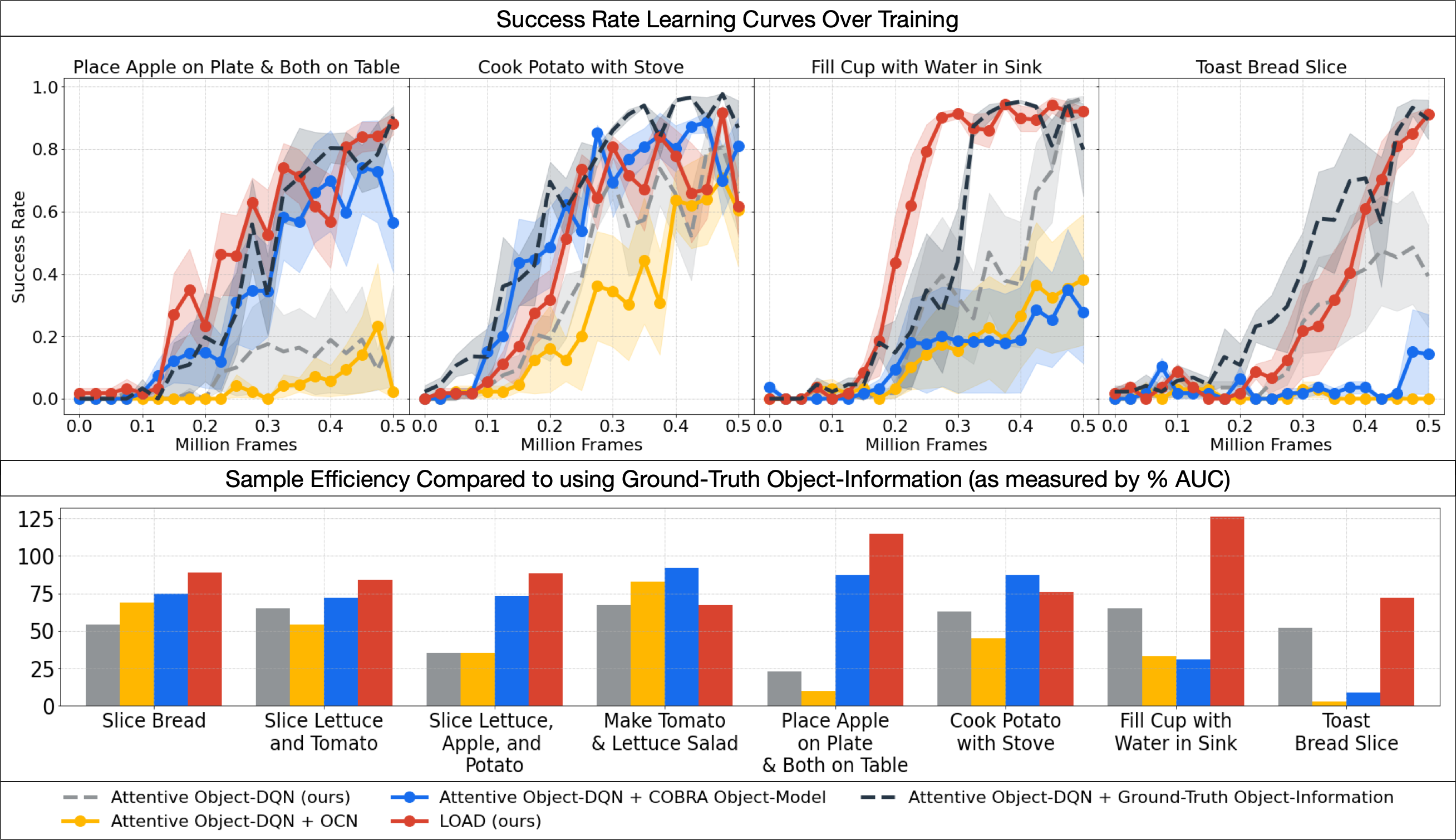}
    \caption{\textbf{Top-panel}: we present the success rate over learning for competing auxiliary tasks.
      We seek a method that best enables our Attentive Object-DQN (grey) to obtain the sample-efficiency it would from adding Ground-Truth Object-Information (black). We visually see that LOAD (red) is best able to learn more quickly on tasks that require using containment-relationships (e.g. a cup in a sink) or recognizing changing object properties (e.g. a toaster turning on with bread in it).
      \newline
      \textbf{Bottom-panel}: by measuring the \% AUC achieved by each agent w.r.t to the agent with ground-truth information, we can measure how close each method is to the performance of an agent with ground-truth object-knowledge. We find LOAD (red), which learns an attentive object-model best closes the performance gap on $6/8$. 
      We hypothesize that this is due to our object-model's ability to capture oracle object-information about object-categories, object-properties, and object-relations. We show evidence for this in Table \ref{table:classification}.
    }
    \label{fig:single_performance}
  \end{minipage}
  \hfill
  \begin{minipage}[b]{0.34\linewidth}
    \centering
    \includegraphics[width=\textwidth]{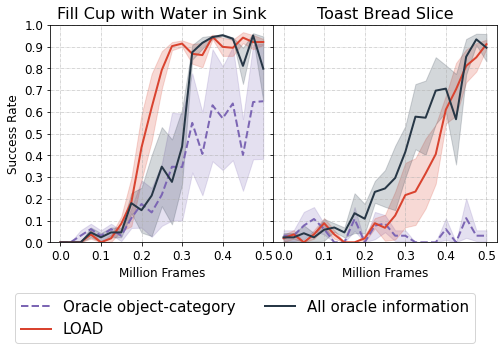}
    \caption{
      Ablation of object-properties and object-relations from oracle. 
      With only oracle object-category information, the oracle can't learn these tasks in our sample budget. 
    }
    \label{fig:object-rep-label}
    \includegraphics[width=\textwidth]{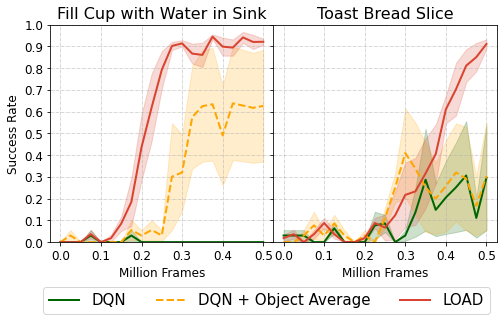}
    \caption{
      Ablation of inter-object attention in policy. Without this, DQN cannot learn these tasks in our sample-budget. See \S \ref{sec:attention-ablation} for details.
    }
    \label{fig:policy-attention-ablation}
  \end{minipage}
  \vfill
\end{figure*}

The primary aim of our experiments is to study how different auxilliary tasks for learning object-representations enable sample complexity comparable to an agent with oracle object-knowledge. We additionally study the degree to which each auxilliary task enables object-representation learning that captures the ground-truth knowledge present in our oracle agent. We conclude this section with ablation experiments studying the importance of different forms of object-knowledge in task learning.

\textbf{Evaluation Settings}.
The agent's spawning location is randomized from $81$ grid positions. The agent receives a terminal reward of $1$ if its task is completed successfully and $0$ otherwise. It receives a time-step penalty of $-0.04$. Episodes have a time-limit of $500$ time-steps. The agent has a budget of $500K$ samples to learn a task. This was the budget needed by a relational agent with oracle object-information.

\newcommand{\noauxtask}{Attentive Object-DQN}

\textbf{Baseline methods for comparison}.
In order to study the effects of competing object representation learning methods, we compare combining Attentive Object-DQN with the Attentive Object-Model against four baseline methods:
\begin{enumerate}[labelindent=0cm,labelsep=2pt,noitemsep,nolistsep, topsep=0pt,leftmargin=*]
  \item \textbf{\noauxtask}. This baseline has no auxiliary task and lets us study how well an agent can learn from the sparse-reward signal alone.
  \item \textbf{Ground-Truth Object-Information}. This baseline has no auxiliary task. Instead, we supply the agent with 14 ground-truth features from the simulator. They roughly describe an object's category (is it a toaster?), its properties (e.g., is it on/off/etc.?), and relevant object-containment (e.g., what object is this object inside of?). Please see \S \ref{appendix:architecture} for detailed descriptions of these features.
  \item \textbf{OCN}. The Object Contrastive Network~\citep{pirk2019online}. This method also employs a classification-like contrastive learning objective to cluster object-images across time-steps. However, it doesn't use an object-model or incorporate action-information. This enables us to study the importance of incoporating an object-model and action information.
  \item \textbf{COBRA Object-Model}. This is the object-model employed by the COBRA RL agent~\citep{watters2019cobra}. They also targeted improved sample-efficiency---though in a simpler, fully-observable 2D environment with shapes that only needed differentiation by category. Their model had no mechanism for incorporating inter-object relations into its predictions. 
\end{enumerate}
To enable faster learning in a sparse-reward setting, all baselines sample training batches using a second self-imitation learning replay buffer of successful episodes~\citep{oh2018self}.

\subsection{Task Performance}
\label{sec:results}
\textbf{Metrics}. We evaluate agent performance by measuring the agent's success rate over 5K frames every 25K frames of experience. The success rate is the proportion of episodes that the agent completes. We compute the mean and standard error of these values across $5$ seeds. To  study sample-efficiency, we compare each method to ``Ground-Truth Object-Information'' by computing what percent of the Ground-Truth Object-Information mean success rate AUC each method achieved. 

We present sample-efficiency bar plots for all 8 of our tasks in  \Figref{fig:single_performance}. We found that using containment relationships and recognizing changing object-properties (Challenges C \& D in \S\ref{sec:framework}) were most indicative of task difficulty. We only present learning curve results for 4 tasks which match this criteria. We present all learning curves in \Figref{fig:single_performance_all} in appendix \S \ref{sec:additional}. We additionally present the maximum success rate achieved by each method in Table \ref{table:success} in appendix \S \ref{sec:success_table}.

\textbf{Performance}. 
We find that using Ground-Truth Object-Information is able to get the highest success rate on all tasks. 
\noauxtask{} performs below all methods besides OCN on $7/8$ tasks. Surprisingly, \noauxtask{} outperforms OCN on $5/8$ tasks. OCN doesn't incorporate action-information when learning to represent object-images across time-steps. We hypothesize that this leads it to learn degenerate object-representations that cannot discriminate object-propertiess that change due to actions, something important for our tasks. 

In terms of sample-efficiency, our Attentive Object-Model comes closest to Ground-Truth Object-Information on $6/8$ tasks. For tasks that require using objects together, such as ``Fill Cup with Water'' where a cup must be used in a sink or ``Toast Bread Slice'' where bread must be cooked in a toaster, our Attentive Object-Model significantly improves over the COBRA Object-Model. 
Interestingly, sample-efficiency goes above 100\% on 2 tasks. We suspect that this is because the object-model provides a learning signal for inter-object attention which is not provided by oracle information.

\subsection{Analysis of Learned Object Representations}
\label{sec:rep-learning}
In \Tableref{table:classification}, we explore our conjecture that the key to strong task-learning performance is an agent's ability to capture the information present in the oracle agent. To study this, we freeze the parameters of each encoding function, and add a linear layer to predict object-categories, object-properties, and containment relationships using a dataset of collected object-interactions we construct (see Appendix \ref{appendix:datasets} for details on the dataset and training). We find that our object-model best captures the information present in the oracle agent.
\begin{table}[!htb]
  \begin{scriptsize}
    \begin{center}
      \input{main/experiments/classification-table.tex}
    \end{center}
    \end{scriptsize}
    \caption{Performance of different unsupervised learning methods for learning object-features (see \S \ref{sec:rep-learning} for details). We find that our object-model best captures features present in the oracle agent, providing evidence that its strong object-representation learning is responsible for its strong task-learning performance.}
  \label{table:classification}
\end{table}

\subsection{Ablations}
\label{sec:properties-relations}
\textbf{Importance of object-properties \& object-relations}.
To verify that capturing object-properties and -relations is key, we train an agent with only oracle object-category information. We find that this agent is not able to learn tasks that require using objects together as object-properties change in our sample-budget (see \Figref{fig:object-rep-label}).

\label{sec:attention-ablation}
\textbf{Importance of inter-object attention}. 
In order to verify the utility of using attention as an inductive bias for capturing object-relations, we ablate attention from both Attentive Object-DQN and our Attentive Object-Model. First, we look at two variants of Attentive Object-DQN without attention. The first is a regular DQN. In the second, we incorporate inter-object information by using the average of all present object-embeddings (DQN + Object Average). Neither learns our tasks in the sample-budget (see \Figref{fig:policy-attention-ablation}).

Additionally, we look at performance where our policy can use inter-object attention but remove inter-object attention from our object-model. Without attention, we still get relatively good performance with 70\% success rate; however, attention in the object-model helps increase this to 90\%+ (see \Figref{fig:model-attention-ablation} in our appendix for details).

%% file: main/experiments/classification-table.tex
\begin{tabular}{l|cccc}
\toprule
                          \twoline{Representation Learning}{Method} & Category & Object-Properties & \twoline{Containment}{Relationship}  \\
\midrule
                            OCN &       $39.2 \pm 8.2$ &    $66.5 \pm 8.5$ &           $69.1 \pm 9.0$ \\
                            COBRA Object-Model &       $79.8 \pm 2.8$ &    $73.4 \pm 8.9$ &           $83.1 \pm 5.8$ \\
  Attentive Object-Model &       $\mathbf{88.6 \pm 3.5}$ &    $\mathbf{98.6 \pm 0.3}$ &           $\mathbf{94.3 \pm 0.6}$ \\
\bottomrule
\end{tabular}

%% file: main/conclusion.tex
We have shown that learning an attentive object-model can enable sample-efficient learning in high-fidelity, 3D, object-interaction domains without access to expert demonstrations or ground-truth object-information. 
Further, when compared to strong unsupervised learning baselines, we have shown that our object-model best captures object-categories, object-properties, and containmennt-relationships. 
We believe that LOAD is a promising steps towards agents that can efficiently learn complex object-interaction tasks.

%% file: supplementary/architectures.tex
\begin{small}
  \begin{table*}[!htb]
    \centering
    \begin{tabular}{l|c} 
      \toprule
      Networks & Parameters \\
      \midrule
      & Attentive Object-DQN \\
      \midrule
      Activation fn. (AF) & Leaky ReLU (LR) \\
      $f_{\tt enc}^{\tt ego}$ & Conv(32-8-4)-AF-Conv(64-4-2)-AF-Conv(64-3-1)-AF-MLP(9216-512)-AF \\
      $f_{\tt enc}^{o}$ & Conv(32-4-2)-AF-Conv(64-4-2)-AF-Conv(64-4-2)-AF-MLP(4096-512)-AF \\
      $f_{\tt enc}^{\tt loc}$ & MLP(6-256)-AF-MLP(256-256)-AF \\
      $\widehat{Q}_{\tt int}(o_i)$ & MLP(1280-256)-AF-MLP(256-256)-AF-MLP(256-8) \\
      $\widehat{Q}_{\tt nav}$ & MLP(768-256)-AF-MLP(256-256)-AF-MLP(256-8) \\
      $\attention(\vz^{o,i}, \mZ^o)$ : $W_1^k$, $W_1^q$ & MLP(512-64), MLP(512-64) \\
      $\attention(\vz^{\kappa}, \mZ^o)$ : $W_2^k$, $W_2^q$ & MLP(768-64), MLP(512-64) \\
      \midrule
      & Object-centric model \\
      \midrule
      $f_{\tt model}$ & MLP(1088-256)-AF-MLP(256-512) \\
      $\vz^a$ : $W^{o}, W^{b}$ & MLP(512-64), MLP(8-64)  \\
      \midrule
      & Scene-centric model \\
      \midrule
      $f^{\kappa}_{\tt model}$ & MLP(832-512) \\
      $\vz^a$ : $W^{o}, W^{b}$ & MLP(512-64), MLP(8-64)  \\
      \midrule
      & VAE \\
      \midrule
      $f_{\tt recon}$ & MLP(4096-512)-AF-Conv(64-4-2)-AF-Conv(64-4-2)-AF-Conv(32-3-2)\\
      \bottomrule
    \end{tabular}
    \vspace{0.1in}
    \caption{Architectures used across all experiments.} \label{table:architectures}
  \end{table*}
\end{small}

We present the details of the architecture used for all models in table \ref{table:architectures}. All models shared the Attentive Object-DQN as their base. We built the Attentive Object-DQN using the \href{https://github.com/vitchyr/rlkit}{rlkit} open-source reinforcement-learning library.

\textbf{Attentive Object-DQN}. This is our base architecture.
Aside from the details in the main text, we note that $f_{\tt enc}^{\kappa}$ is the concatenation of one function which encodes image information and another that encodes location information: $ f_{\tt enc}^{\kappa}(s^{\kappa}) = f_{\tt enc}^{\kappa}(s^{\tt ego}, s^{\tt loc})  = [f_{\tt enc}^{\tt ego}(s^{\tt ego}), f_{\tt enc}^{\tt loc}(s^{\tt loc})]$.

\textbf{Attentive Object-DQN + COBRA Object-Model}: each agent predicts the latent factors that have generated each individual object-image-patch. This requires an additional reconstruction network for the object-encoder, $f_{\tt recon}(\vz_t^{o,i})$, which produces an object-image-patch back from an encoding and a prediction network $f_{\tt cobramodel}$ that produces the object-encoding for $\vz_t^{o,i}$ at the next time-step. The objective function is:
\begin{equation}
  \begin{split}
  \mathcal{L}_{\tt recon} 
  &= \E_{s_t} \left[ \sum_{i\in v_{t}} ||f_{\tt recon}(\vz_t^{o,i}) - o_{t,i} ||^2_2\right] \\
  & - \E_{s_t} \left[ \beta^{\tt kl}\operatorname{KL}(p(\vz_t^{o,i}|o_{t,i})|| p(\vz_t^{o,i})) \right] 
  \end{split}
\end{equation}
\begin{equation}
  \begin{split}  
  \mathcal{L}_{\tt pred} 
  &= \E_{s_t} \left[ \sum_{i\in v_{t}} ||f_{\tt recon}(f_{\tt cobramodel}(\vz_t^{o,i})) - o_{t+1,i} ||^2_2 \right] \\
  \end{split}
\end{equation}
\begin{equation}
  \begin{split}
\mathcal{L}_{\tt cobra} 
  &= \mathcal{L}_{\tt recon}  + \mathcal{L}_{\tt pred}  \\
  & - \E_{s_t} \left[ \sum_{i\in v_{t}} \beta^{\tt kl}\operatorname{KL}(p(\vz_t^{o,i}|o_{t,i})|| p(\vz_t^{o,i})) \right] 
  \end{split}
\end{equation}
where $\operatorname{KL}$ is the Kullback-Leibler Divergence and $p(\vz_t^{o,i})$ is an isotropic, unit gaussian. We also model $p(\vz_t^{o,i}|o_{t,i})$ as a gaussian. We augment the Attentive Object-DQN so that $\vz_t^{o,i}$ is the mean of the gaussian and so that a standard deviation is also computed. Please see \citet{higgins2017beta} for more.

\textbf{Attentive Object-DQN + OCN}: the agent tries to learn encodings of object-image-patches such that patches across time-steps corresponding to the same object are grouped nearby in latent space, and patches corresponding to different objects are pushed apart. This also relied on contrastive learning, except that it uses it on image-pairs across time-steps.
Following \citet{pirk2019online}, the anchor is defined as the object-encoding $\vz_t^{o,i}=f(o_{t,i})$, which we will refer to as $f$. 
The positive is defined as the object-image-patch encoding at the next time-step with lowest $L2$ distance in latent space, $f^{+} = \argmin_{z_{t+1}^{o,j}} ||\vz_{t}^{o,i} - \vz_{t+1}^{o,j} ||^2$. We then set negatives $\{f_k^{-}\}$ as the object-image-patches that did not correspond to the match. We note that augmenting \citet{pirk2019online} so that their objective function had temperature $\tau$ was required for good performance. For a unified perspective with our own objective function, we write their n-tuplet-loss with a softmax (see \citet{Sohn2016improved_metric} for more details on their equivalence). The objective function is:
\begin{align}
  \mathcal{L}_{\tt ocn} 
   = \E\left[
    -\log \left(\frac{\exp \left( f^{\top} f^+/\tau\right) }{ \exp \left(f^{\top} f^+/\tau \right) + \sum_k \exp \left(f^{\top} f_k^{-}/\tau )\right) } \right)  
    \right]
\end{align}

\textbf{Attentive Object-DQN + Ground Truth Object Info}: the agent doesn't have an auxiliary task and doesn't encode object-images. Instead it encodes object-information. For each object, we replace object-image-patches with the following information available in Thor:
\begin{enumerate}
  \item object-category. If the object is a toaster, this would be the index corresponding to toaster.
\end{enumerate}
The following correspond to ``object-relations'':
\begin{enumerate}
\item What object is this object inside of (e.g. if this object is a cup in the sink, this would correspond to the sink index).
\item What object is inside this object of (e.g. if this object is a sink with a cup in it, this would correspond to the cup index).
\end{enumerate}
The following correspond to ``object-properties'':
\begin{enumerate}
  \item distance to object (in meters)
  \item whether object is visible (boolean)
  \item whether object is toggled (boolean)
  \item whether object is broken (boolean)
  \item whether object is filledWithLiquid (boolean)
  \item whether object is dirty (boolean)
  \item whether object is cooked (boolean)
  \item whether object is sliced (boolean)
  \item whether object is open (boolean)
  \item whether object is pickedUp (boolean)
  \item object temperature (cold, room-temperature, hot)
\end{enumerate}

%% file: supplementary/hyperparameter.tex
\begin{table*}[!htb]
  \centering
  \begin{tabular}{l|c|c} 
  \toprule
  Hyperparameter & Final Value & Values Considered \\
  \midrule

  Max gradient norm & $0.076$ & $\operatorname{log-uniform}(10^{-4}, 10^{-1})$ \\
  \midrule\midrule
  \textbf{DQN} & & \\
  \midrule
  Learning rate $\eta_1$ & $1.8 \times 10^{-5}$ & $\operatorname{log-uniform}(10^{-6}, 10^{-2})$ \\
  Target Smoothing Coeffecient $\eta_2$ & $0.00067$ & $\operatorname{log-uniform}(10^{-6}, 10^{-3})$ \\
  Discount $\gamma$ & $0.99$ & \\
  Training $\epsilon$ annealing  & $[1, .1]$ & \\
  Evaluation $\epsilon$ & $.1$ & \\
  Regular Replay Buffer Size & $150000$ & \\
  SIL Replay Buffer Size & $50000$ & \\
  Regular:SIL Replay Ratio & $7:1$ & \\
  Batchsize & $50$ & \\
  \midrule\midrule

  \textbf{Attentive Object-Model} & & \\
  \midrule
  upper-bound $m$ & $85$ & - \\
  Number of Negative Examples & $20$ & - \\
  temperature $\tau$ & $8.75 \times 10^{-5}$ & $\operatorname{log-uniform}(10^{-6}, 10^{-3})$ \\
  Loss Coeffecient $\beta^{\tt model}$ & $10^{-3}$ & - \\
  \midrule\midrule

  \textbf{Cobra Object-Model} & & \\
  \midrule
  KL Coeffecient $\beta^{\tt kl}$ & $26$ & $\operatorname{log-uniform}(10^{-1}, 10^{2})$ \\
  Loss Coeffecient $\beta^{\tt cobra}$ & $0.0032$ & $\operatorname{log-uniform}(10^{-4}, 1)$ \\
  \midrule\midrule

  \textbf{OCN} & & \\
  \midrule
  temperature $\tau$ & $5 \times 10^{-5}$ & $\operatorname{log-uniform}(10^{-6}, 10^{-3})$ \\
  Loss Coeffecient $\beta^{\tt ocn}$ & $0.0047$ & $\operatorname{log-uniform}(10^{-4}, 10^{-2})$ \\

  \bottomrule
  \end{tabular}
  \vspace{0.1in}
  \caption{Hyperparameters shared across all experiments.} \label{table:hyperparameters}
\end{table*}

\textbf{Attentive Object-DQN}.
All models are based on the same Attentive Object-DQN agent and thus use the same hyperparameters. We searched over these parameters using ``Attentive Object-DQN + Ground-Truth Object-Information''. We searched over tuples of the parameters in the ``DQN'' portion of table \ref{table:hyperparameters}. In addition to searching over those parameters, we searched over ``depths'' and hidden layer size of the multi-layer perceptrons $f_{\tt enc}^{\tt loc}$, $\widehat{Q}_{\tt int}(o_i)$, and $\widehat{Q}_{\tt nav}$. For depths, we searched uniformly over $[0,1,2]$ and for hidden later sizes we searched uniformly over $[128, 256, 512]$. We searched over 12 tuples on the ``Fill Cup with Water'' task and 20 tuples on the ``Place Apple on Plate \& Both on Table'' task. We found that task-performance was sensitive to hyperparameters and choose hyperparameters that achieved a $90\%+$ success rate on both tasks. We fixed these settings and searched over the remaining values for each auxiliary task.

\textbf{Attentive Object-Model}. 
We experimented with the number of negative examples used for the contrastive loss and found no change in performance. We performed a search over 4 tuples from the values in table \ref{table:hyperparameters}. We chose the loss-coefficient as the the coefficient which put the object-centric model loss at the same order of magnitude as the DQN loss.

\textbf{COBRA Object-Model, OCN}. 
For each auxiliary task, we performed a search over 6 tuples from the values in table \ref{table:hyperparameters}. For each loss, we chose loss coefficients that scaled the loss so they were between an order of magnitude above and below the DQN loss.

%% file: supplementary/thor-settings.tex
\textbf{Environment}. 
While AI2Thor has multiple maps to choose from, we chose ``Floorplan 24''.
To reduce the action-space, we restricted the number of object-types an agent could interact with so that there were $10$ distractor types beyond task relevant object-types. We defined task-relevant object-types as objects needed to complete the task or objects they were on/inside. For example, in ``Place Apple on Plate \& Both on Table'', since the plate is on a counter, counters are task object-types. We provide a list of the object-types present in each task with the task descriptions below.

\textbf{Observation}. Each agent observes an $84 \times 84$ grayscale image of the environment, downsampled from a $300 \times 300$ RGB image. They can detect up to $20$ obects per time-step within its line of sight, if they exceed $50$ pixels in area, regardless of distance. Each object in the original $300 \times 300$ scene image is cropped and resized to a $32 \times 32$ grayscale image\footnote{For ``Slice'' tasks and ``Make Tomato \& Lettuce Salad'', we used an object image size of $64 \times 64$ to facilitate recognition of smaller objects. We decreased the replay buffer to have $90000$ samples and the SIL replay buffer to have $30000$ samples. }. Each agent observes its $(x,y,z)$ location, and its pitch, yaw, and roll body rotation $(\varphi_1, \varphi_2, \varphi_3)$ in a global coordinate frame.

\textbf{Episodes}. The episode terminates either after $500$ steps or when a task is complete. The agent's spawning location is randomly sampled from the $81$ grid positions facing North with a body angle $(0^{\circ}, 0^{\circ}, 0^{\circ})$. Each agent recieves reward $1$ if a task is completed successfully and a time-step penalty of $-0.04$. 

\begin{small}
\begin{table}[!htb] 
  \centering
    \begin{tabular}{l|c} 
      \toprule
      Setting & Values \\
      \midrule
      Observation Size & $300 \times 300$ \\
      Downsampled Observation Size & $84 \times 84$ \\
      Object Image Size & $32 \times 32$ \\
      Min Bounding Box Proportion & $\frac{50}{300 \times 300}$ \\
      Max Interaction Distance & $1.5$m \\
      \bottomrule
    \end{tabular}  
    \vspace{0.1in}
    \caption{Settings used in Thor across experiments.} \label{table:thor-settings}
  \end{table}  
\end{small}  

%% file: supplementary/task_details.tex
For each task, we describe which challenges were present, what object types were interactable, and the total Key Semantic Actions available. We chose objects that were evenly spaced around the environment. The challenges were:
\begin{description}
  \item[Challenge A:] the need for view-invariance (e.g. recognizing a knife across
  angles),
  \item[Challenge B:] the need to reason over $\geq 3$ objects,
  \item[Challenge C:] the need to recognize and use \textit{combined} objects (e.g. filling a cup with water in the sink or toasting bread in a toaster).
\end{description}

\textbf{Slice Bread}.
\\\textit{Challenges}:
\begin{enumerate}[noitemsep]
  \item[A:] recognizing the knife across angles.
\end{enumerate}
\textit{Interactable Object Types}: 15
\begin{itemize}[noitemsep]
  \item CounterTop: 3, DiningTable: 1, Microwave: 1, Plate: 1, CoffeeMachine: 1, Bread: 1, Fridge: 1, Egg: 1, Cup: 1, Pot: 1, Pan: 1, Tomato: 1, Knife: 1
\end{itemize}
\textit{Key Semantic Actions}: 
\begin{enumerate}[noitemsep]
  \item Go to Knife
  \item Pickup Knife
  \item Go to Bread
  \item Slice Bread
\end{enumerate}

\textbf{Slice Lettuce and Tomato}. (order doesn't matter)
\\\textit{Challenges}:
\begin{enumerate}[noitemsep]
  \item[A:] recognizing the knife across angles.
  \item[B:] recognizing and differentiate 3 task objects: the knife, lettuce, and tomato. As each object is cut, the agent needs to choose from more objects as it can select from the object-slices. 
\end{enumerate}
\textit{Interactable Object Types}: 17
\begin{itemize}[noitemsep]
  \item CounterTop: 3, DiningTable: 1, Microwave: 1, Plate: 1, CoffeeMachine: 1, Bread: 1, Fridge: 1, Spatula: 1, Egg: 1, Cup: 1, Pot: 1, Pan: 1, Tomato: 1, Lettuce: 1, Knife: 1
\end{itemize}
\textit{Key Semantic Actions}:
\begin{enumerate}[noitemsep]
  \item Go to Knife
  \item Pickup Knife
  \item Go to Table
  \item Slice Lettuce
  \item Slice Tomato
\end{enumerate}

\textbf{Slice Lettuce and Apple, and Potato}. (order doesn't matter)
\\\textit{Challenges}:
\begin{enumerate}[noitemsep]
  \item[A:] recognizing the knife across angles.
  \item[B:] recognizing and differentiate 4 task objects: the knife, lettuce, and apple, and potato. As each object is cut, the agent needs to choose from more objects as it can select from the object-slices.
\end{enumerate}
\textit{Interactable Object Types}: 18
\begin{itemize}[noitemsep]
  \item CounterTop: 3, DiningTable: 1, Microwave: 1, Plate: 1, CoffeeMachine: 1, Bread: 1, Fridge: 1, Potato: 1, Egg: 1, Cup: 1, Pot: 1, Pan: 1, Tomato: 1, Lettuce: 1, Apple: 1, Knife: 1
\end{itemize}
\textit{Key Semantic Actions}:
\begin{enumerate}[noitemsep]
  \item Go to Knife
  \item Pickup Knife
  \item Go to Table
  \item Slice Lettuce
  \item Slice Apple
  \item Slice Potato
\end{enumerate}

\textbf{Cook Potato on Stove}. 
\\\textit{Challenges}:
\begin{enumerate}[noitemsep]
  \item[A:] recognizing the stove across angles. 
  \item[B:] needs to differentiate 3 objects: the stove knob, pot, and potato.
  \item[C:] recognizing the potato in the pot.
\end{enumerate}
\textit{Interactable Object Types}: 21
\begin{enumerate}[noitemsep]
  \item StoveBurner: 4, StoveKnob: 4, DiningTable: 1, Microwave: 1, Plate: 1, CoffeeMachine: 1, Bread: 1, Fridge: 1, Potato: 1, Egg: 1, Cup: 1, Pot: 1, Pan: 1, Tomato: 1, Knife: 1
\end{enumerate}
\textit{Key Semantic Actions}:
\begin{enumerate}[noitemsep]
  \item Go to Potato
  \item Pickup Potato
  \item Go to Stove
  \item Put Potato in Pot
  \item Turn on Stove Knob
\end{enumerate}

\textbf{Fill Cup with Water}. 
\\\textit{Challenges}:
\begin{enumerate}[noitemsep]
  \item[A:] recognizing the cup across angles and backgrounds. 
  \item[B:] recognizing the cup in the sink.
  \item[C:] the need to recognize and use \textit{combined} objects (e.g. filling a cup with water in the sink or toasting bread in a toaster).
\end{enumerate}
\textit{Interactable Object Types}: 18
\begin{itemize}[noitemsep]
  \item CounterTop: 3, Faucet: 2, Sink: 1, DiningTable: 1, Microwave: 1, CoffeeMachine: 1, Bread: 1, Fridge: 1, Egg: 1, Cup: 1, SinkBasin: 1, Pot: 1, Pan: 1, Tomato: 1, Knife: 1
\end{itemize}
\textit{Key Semantic Actions}: 
\begin{enumerate}[noitemsep]
  \item Go to Cup
  \item Pickup Cup
  \item Go to Sink
  \item Put Cup in Sink
  \item Fill Cup
\end{enumerate}

\textbf{Toast Bread Slice}.
\\\textit{Challenges}:
\begin{enumerate}[noitemsep]
  \item[A:] recognizing the toaster across angles.
  \item[C:] recognizing the bread slice in the toaster.
\end{enumerate}
\textit{Interactable Object Types}: 21
\begin{itemize}[noitemsep]
  \item BreadSliced: 5, CounterTop: 3, Bread: 2, DiningTable: 1, Microwave: 1, CoffeeMachine: 1, Fridge: 1, Egg: 1, Cup: 1, Pot: 1, Pan: 1, Tomato: 1, Knife: 1, Toaster: 1
\end{itemize}
\textit{Key Semantic Actions}:  
\begin{enumerate}[noitemsep]
  \item Go to Bread Slice
  \item Pickup Bread Slice
  \item Go to Toaster
  \item Put Breadslice in Toaster
  \item Turn on Toaster
\end{enumerate}

\textbf{Place Apple on Plate \& Both on table}.
\\\textit{Challenges}:
\begin{enumerate}[noitemsep]
  \item[B:] needs to differentiate 3 objects: the apple, plate, and table.
  \item[C:] recognizing the apple on the plate.
\end{enumerate}
\textit{Interactable Object Types}: 16
\begin{itemize}[noitemsep]
  \item CounterTop: 3, DiningTable: 1, Microwave: 1, Plate: 1, CoffeeMachine: 1, Bread: 1, Fridge: 1, Spatula: 1, Egg: 1, Cup: 1, Pot: 1, Pan: 1, Apple: 1, Knife: 1
\end{itemize}
\textit{Key Semantic Actions}:
\begin{enumerate}[noitemsep]
  \item Go to Apple
  \item Pickup Apple
  \item Put Apple on Plate
  \item Pickup Plate
  \item Go to Table
  \item Put Plate on Table
\end{enumerate}

\textbf{Make Tomato \& Lettuce Salad}.
\\\textit{Challenges}:
\begin{enumerate}[noitemsep]
  \item[B:] needs to differentiate 3 objects: the tomato slice, lettuce slice, and plate.
  \item[C:] recognizing the tomato slice or lettuce slice on the plate.
\end{enumerate}
\textit{Interactable Object Types}: 32
\begin{itemize}[noitemsep]
  \item TomatoSliced: 7, LettuceSliced: 7, CounterTop: 3, Bread: 2, DiningTable: 1, Microwave: 1, Plate: 1, CoffeeMachine: 1, Fridge: 1, Spatula: 1, Egg: 1, Cup: 1, Pot: 1, Pan: 1, Tomato: 1, Lettuce: 1, Knife: 1
\end{itemize}
\textit{Key Semantic Actions}:
\begin{enumerate}[noitemsep]
  \item Go to table 
  \item Pickup tomato or lettuce slice
  \item Put slice on Plate
  \item Pickup other slice
  \item Put slice on Plate
\end{enumerate}

%% file: supplementary/datasets.tex
\label{appendix:datasets}
In order to measure and analyze the quality of the object representations learned via each auxiliary task, we created a dataset with programmatically generated object-interactions and with random object-interactions. This enabled us to have a diverse range of object-interactions and ensured the dataset had many object-states present.

\textbf{Programatically Generated object-interactions}. This dataset contains programmatically generated sequences of interactions for various tasks. The tasks currently supported by the dataset include: \textit{pickup X, turnon X, open X, fill X with Y, place X in Y, slice X with Y, Cook X in Y on Z.} For each abstract task type, we first enumerate all possible manifestations based on the action and object properties. For example, manifestations of open X include all objects that are openable. We exhaustively test each manifestation and identify the ones that are possible under the physics of the environment. We explicitly build the action sequence required to complete each task. Because we only want to collect object-interactions, we use the high level “TeleportFull” command for navigation to task objects. The TeleportFull command allows each agent to conveniently navigate to desired task objects at a particular location and viewing angle. For example, the sequence for place X in Y is: TeleportFull to X, Pickup X, TeleportFull to Y, and Put X in Y. An agent will execute each action until termination. We collect both successful and unsuccessful task sequences. There is a total of 156 unique tasks in the dataset and 1196 individual task sequences amounting to 2353 $(\text{state}, \text{action}, \text{next state})$ tuples.

\textbf{Random object-interactions}. The random interaction dataset consists of $(\text{state}, \text{action}, \text{next state})$ tuples of random interactions with the environment. 
An agent equipped with a random action policy interacts with the environment for episodes of 500 steps until it collects a total of 4000 interaction samples.

\textbf{Features in dataset}. 
We study the following features: \textit{Category} is a multi-class label indicating an object's category. The following are binary labels. 
\textit{Object-properties} contains 6 features such as whether objects \textit{are} closed, turned on, etc. \textit{Containment Relationship} contains 2 featues: whether an object is inside another object or whether another object inside of it. For each feature-set, we present the mean average precision and standard error for each method across all 8 tasks in Table \ref{table:classification}. 

\textbf{Training}. We divided the data into an 80/20 training/evaluation split and trained for 2000 epochs. We reported the test data results.

%% file: supplementary/additional_results.tex
\begin{figure*}[b]
  \centering
  \includegraphics[width=0.8\textwidth]{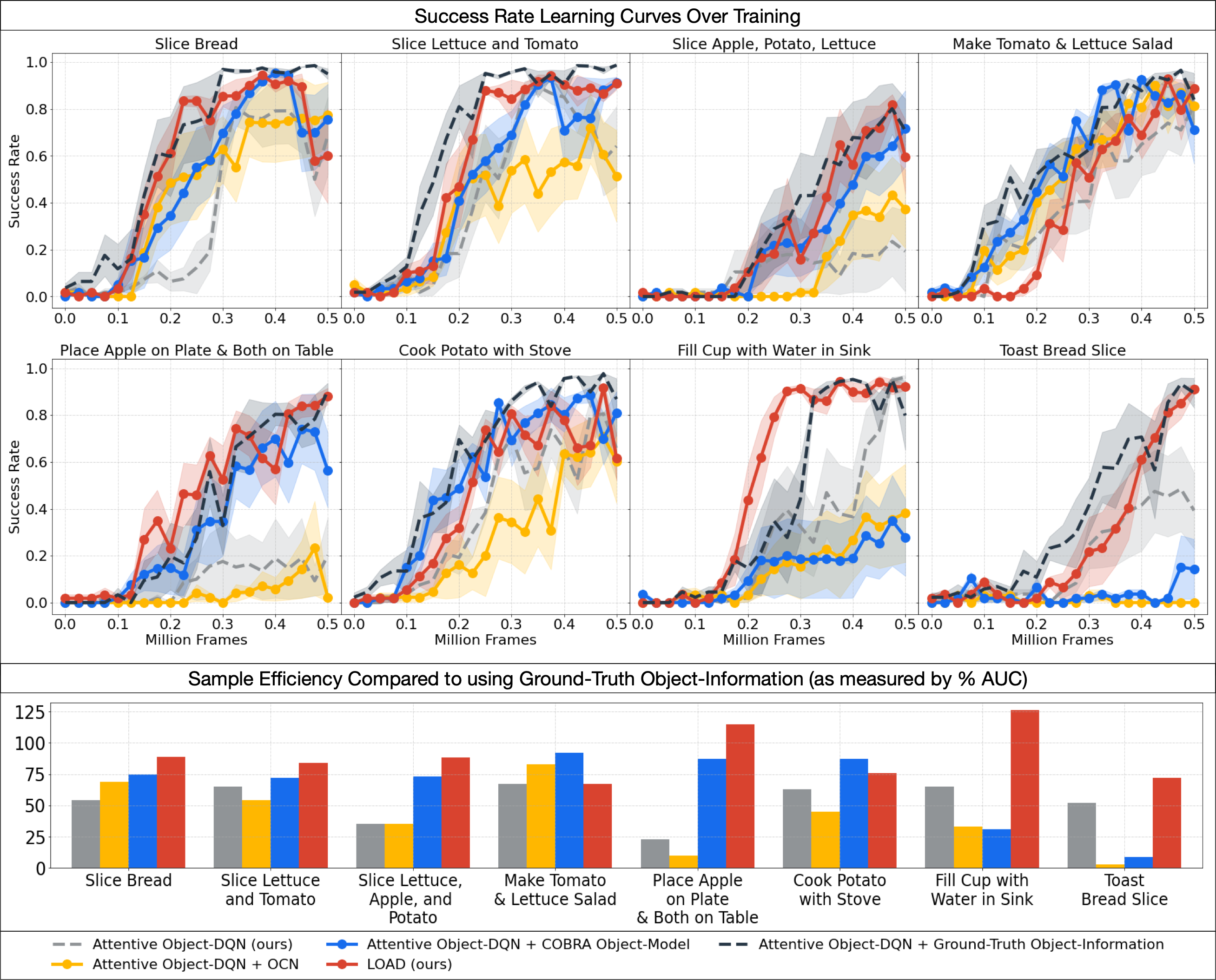}
    \caption{\textbf{Top-panel}: we present the success rate over learning for competing auxiliary tasks.
    We seek a method that best enables our Attentive Object-DQN (grey) to obtain the sample-efficiency it would from adding Ground-Truth Object-Information (black). We visually see that LOAD (red) is best able to learn more quickly on tasks that require using containment-relationships (e.g. a cup in a sink) or recognizing changing object properties (e.g. a toaster turning on with bread in it).
    \newline
    \textbf{Bottom-panel}: by measuring the \% AUC achieved by each agent w.r.t to the agent with ground-truth information, we can measure how close each method is to the performance of an agent with ground-truth object-knowledge. We find LOAD (red), which learns an attentive object-model best closes the performance gap on $6/8$. 
    We hypothesize that this is due to our object-model's ability to capture oracle object-information about object-categories, object-properties, and object-relations. We show evidence for this in Table \ref{table:classification}.
  }
  \label{fig:single_performance_all}
\end{figure*}

\subsection{Success rate of competing auxiliary tasks}
\label{sec:success_table}
\begin{table*}[!htb]
  \begin{center}
    \input{main/experiments/success.tex}
    \caption{Maximum success rate achieved by competing auxiliary tasks during training.}
    \label{table:success}
  \end{center}
\end{table*}
To supplement the training success curves in \S \ref{sec:results}, we also provide the maximum success rates obtained by each auxiliary task in \Tableref{table:success}. In \Tableref{table:success}, we find that using Ground-Truth Object-Information is able to get the highest success rate on $7/8$ tasks. It only achieves 80\% on ``Slice Apple, Potato, Lettuce'', a task that requires using 4 objects, which is consistent with our finding that tasks that require more objects have a higher sample-complexity.

In terms of maximum success rate, looking at \Tableref{table:success}, our Attentive Object-Model comes closest to Ground-Truth Object-Information on $5/8$ tasks and is tied on $3/8$ tasks with the COBRA Object-Model. However, for tasks that require using objects together, such as ``Fill Cup with Water'' where a cup must be used in a sink or ``Toast Bread Slice'' where bread must be cooked in a toaster, the COBRA Object-Model exhibits a higher sample-complexity.

\subsection{Ablating inter-object attention from Attentive Object-Model}
\begin{figure}[!htb]
  \centering
  \includegraphics[width=0.48\textwidth]{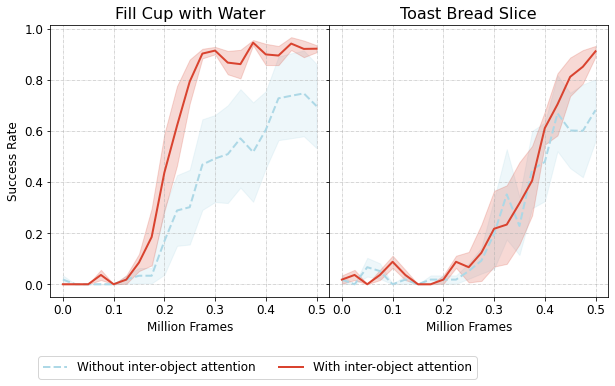}
  \caption{Ablation of inter-object attention in object-model. We show  that incorporating inter-object attention into our object-model leads to better perfomance.}
  \label{fig:model-attention-ablation}
\end{figure}

We ablate inter-object attention from our agent's model. In \Figref{fig:model-attention-ablation}, we find that the agent can perform reasonably well without incorporating attention into the object-model, achieving a success rate of about 70\%. With attention however, the agent can get above a 90\% success rate.

%% file: main/experiments/success.tex
\resizebox{\textwidth}{!}{%

\begin{tabular}{c|cccccccc}
  \toprule
  \twoline{Auxiliary}{Task}&    Slice Bread & \twoline{Slice Lettuce}{and Tomato} & \twoline{Slice Apple,}{Potato, Lettuce} & \twoline{Cook Potato}{on Stove} & \twoline{Fill Cup}{with Water} & \twoline{Toast}{Bread Slice} & \twoline{Apple on Plate,}{ Both on Table} & \twoline{Make}{Salad} \\
  \midrule
  \midrule
  \twoline{No Auxiliary}{Task} &   $80.6 \pm 7.8$ &             $89.6 \pm 3.0$ &                $23.5 \pm 14.7$ &          $80.6 \pm 12.9$ &                $\mathbf{96.3 \pm 0.7}$ &     $48.5 \pm 18.2$ &                        $20.2 \pm 16.1$ &                $81.4 \pm 7.3$ \\
    \hline
    \twoline{Object Contrastive}{Network (OCN)} &  $77.6 \pm 13.9$ &            $72.0 \pm 15.1$ &                $43.4 \pm 16.3$ &           $70.9 \pm 9.6$ &               $38.2 \pm 20.9$ &       $3.0 \pm 1.9$ &                        $23.2 \pm 18.0$ &                $90.2 \pm 1.8$ \\
    \hline
    \twoline{COBRA}{Object-Model} &   $\mathbf{95.3 \pm 1.2}$ &             $\mathbf{93.4 \pm 1.4}$ &                $71.7 \pm 16.2$ &           $88.6 \pm 3.3$ &               $35.0 \pm 19.3$ &     $15.1 \pm 13.5$ &                        $74.1 \pm 14.8$ &                $\mathbf{92.7 \pm 1.4}$ \\
    \hline
    \twoline{Attentive}{Object-Model} &   $\mathbf{94.4 \pm 1.8}$ &             $\mathbf{94.2 \pm 0.5}$ &                 $\mathbf{81.9 \pm 4.4}$ &           $\mathbf{91.7 \pm 2.3}$ &                $\mathbf{94.5 \pm 1.0}$ &      $\mathbf{91.1 \pm 2.1}$ &                         $\mathbf{88.1 \pm 3.4}$ &                $\mathbf{92.8 \pm 0.5}$ \\
    \midrule
    \midrule
    \twoline{Ground-Truth}{Object-Info} &   $\mathbf{98.6 \pm 0.2}$ &             $\mathbf{98.8 \pm 0.2}$ &                $\mathbf{80.2 \pm 10.8}$ &           $\mathbf{97.7 \pm 0.2}$ &                $\mathbf{95.2 \pm 0.4}$ &      $\mathbf{93.3 \pm 2.3}$ &                         $\mathbf{90.5 \pm 3.2}$ &                $\mathbf{96.6 \pm 0.2}$ \\
  \bottomrule
  \end{tabular}

}

%% file: ijcai21-multiauthor.bbl
\begin{thebibliography}{}

\bibitem[\protect\citeauthoryear{Gordon \bgroup \em et al.\egroup
  }{2018}]{gordon2018iqa}
Daniel Gordon, Aniruddha Kembhavi, Mohammad Rastegari, Joseph Redmon, Dieter
  Fox, and Ali Farhadi.
\newblock Iqa: Visual question answering in interactive environments.
\newblock In {\em Proceedings of the IEEE Conference on Computer Vision and
  Pattern Recognition}, 2018.

\bibitem[\protect\citeauthoryear{Hadsell \bgroup \em et al.\egroup
  }{2006}]{hadsell2006dimensionality}
Raia Hadsell, Sumit Chopra, and Yann LeCun.
\newblock Dimensionality reduction by learning an invariant mapping.
\newblock In {\em 2006 IEEE Computer Society Conference on Computer Vision and
  Pattern Recognition (CVPR'06)}, volume~2, pages 1735--1742. IEEE, 2006.

\bibitem[\protect\citeauthoryear{Higgins \bgroup \em et al.\egroup
  }{2017}]{higgins2017beta}
Irina Higgins, Loic Matthey, Arka Pal, Christopher Burgess, Xavier Glorot,
  Matthew Botvinick, Shakir Mohamed, and Alexander Lerchner.
\newblock beta-vae: Learning basic visual concepts with a constrained
  variational framework.
\newblock {\em International Conference on Learning Representations}, 2017.

\bibitem[\protect\citeauthoryear{Jain \bgroup \em et al.\egroup
  }{2019}]{jain2019two}
Unnat Jain, Luca Weihs, Eric Kolve, Mohammad Rastegari, Svetlana Lazebnik, Ali
  Farhadi, Alexander~G Schwing, and Aniruddha Kembhavi.
\newblock Two body problem: Collaborative visual task completion.
\newblock In {\em Proceedings of the IEEE Conference on Computer Vision and
  Pattern Recognition}, 2019.

\bibitem[\protect\citeauthoryear{Kipf \bgroup \em et al.\egroup
  }{2019}]{kipf2019contrastive}
Thomas Kipf, Elise van~der Pol, and Max Welling.
\newblock Contrastive learning of structured world models.
\newblock {\em arXiv preprint arXiv:1911.12247}, 2019.

\bibitem[\protect\citeauthoryear{Kolve \bgroup \em et al.\egroup
  }{2017}]{kolve2017ai2thor}
Eric Kolve, Roozbeh Mottaghi, Daniel Gordon, Yuke Zhu, Abhinav Gupta, and Ali
  Farhadi.
\newblock Ai2-thor: An interactive 3d environment for visual ai.
\newblock {\em arXiv preprint arXiv:1712.05474}, 2017.

\bibitem[\protect\citeauthoryear{Mnih \bgroup \em et al.\egroup
  }{2015}]{mnih2015human}
Volodymyr Mnih, Koray Kavukcuoglu, David Silver, Andrei~A Rusu, Joel Veness,
  Marc~G Bellemare, Alex Graves, Martin Riedmiller, Andreas~K Fidjeland, Georg
  Ostrovski, et~al.
\newblock Human-level control through deep reinforcement learning.
\newblock {\em Nature}, 518(7540):529, 2015.

\bibitem[\protect\citeauthoryear{Oh \bgroup \em et al.\egroup
  }{2015}]{oh2015action}
Junhyuk Oh, Xiaoxiao Guo, Honglak Lee, Richard~L Lewis, and Satinder Singh.
\newblock Action-conditional video prediction using deep networks in atari
  games.
\newblock In {\em Advances in neural information processing systems}, 2015.

\bibitem[\protect\citeauthoryear{Oh \bgroup \em et al.\egroup
  }{2017}]{oh2017zero}
Junhyuk Oh, Satinder Singh, Honglak Lee, and Pushmeet Kohli.
\newblock Zero-shot task generalization with multi-task deep reinforcement
  learning.
\newblock In {\em International Conference on Machine Learning}, 2017.

\bibitem[\protect\citeauthoryear{Oh \bgroup \em et al.\egroup
  }{2018}]{oh2018self}
Junhyuk Oh, Yijie Guo, Satinder Singh, and Honglak Lee.
\newblock Self-imitation learning.
\newblock {\em arXiv preprint arXiv:1806.05635}, 2018.

\bibitem[\protect\citeauthoryear{Pirk \bgroup \em et al.\egroup
  }{2019}]{pirk2019online}
S{\"o}ren Pirk, Mohi Khansari, Yunfei Bai, Corey Lynch, and Pierre Sermanet.
\newblock Online object representations with contrastive learning.
\newblock {\em arXiv preprint arXiv:1906.04312}, 2019.

\bibitem[\protect\citeauthoryear{Reed \bgroup \em et al.\egroup
  }{2014}]{reed2014learning}
Scott Reed, Kihyuk Sohn, Yuting Zhang, and Honglak Lee.
\newblock Learning to disentangle factors of variation with manifold
  interaction.
\newblock In {\em International Conference on Machine Learning}, 2014.

\bibitem[\protect\citeauthoryear{Shridhar \bgroup \em et al.\egroup
  }{2019}]{Shridhar2019ALFREDAB}
Mohit Shridhar, Jesse Thomason, Daniel Gordon, Yonatan Bisk, Winson Han,
  Roozbeh Mottaghi, Luke Zettlemoyer, and Dieter Fox.
\newblock Alfred: A benchmark for interpreting grounded instructions for
  everyday tasks.
\newblock {\em ArXiv}, abs/1912.01734, 2019.

\bibitem[\protect\citeauthoryear{Sohn}{2016}]{Sohn2016improved_metric}
Kihyuk Sohn.
\newblock Improved deep metric learning with multi-class n-pair loss objective.
\newblock In {\em Advances in Neural Information Processing Systems}, 2016.

\bibitem[\protect\citeauthoryear{Van~Hasselt \bgroup \em et al.\egroup
  }{2016}]{van2016deep}
Hado Van~Hasselt, Arthur Guez, and David Silver.
\newblock Deep reinforcement learning with double q-learning.
\newblock In {\em AAAI conference on artificial intelligence}, 2016.

\bibitem[\protect\citeauthoryear{Van~Hoof \bgroup \em et al.\egroup
  }{2015}]{van2015learning}
Herke Van~Hoof, Tucker Hermans, Gerhard Neumann, and Jan Peters.
\newblock Learning robot in-hand manipulation with tactile features.
\newblock In {\em International Conference on Humanoid Robots}, 2015.

\bibitem[\protect\citeauthoryear{Vaswani \bgroup \em et al.\egroup
  }{2017}]{vaswani2017attention}
Ashish Vaswani, Noam Shazeer, Niki Parmar, Jakob Uszkoreit, Llion Jones,
  Aidan~N Gomez, {\L}ukasz Kaiser, and Illia Polosukhin.
\newblock Attention is all you need.
\newblock In {\em Advances in neural information processing systems}, 2017.

\bibitem[\protect\citeauthoryear{Veerapaneni \bgroup \em et al.\egroup
  }{2020}]{veerapaneni2020entity}
Rishi Veerapaneni, John~D Co-Reyes, Michael Chang, Michael Janner, Chelsea
  Finn, Jiajun Wu, Joshua Tenenbaum, and Sergey Levine.
\newblock Entity abstraction in visual model-based reinforcement learning.
\newblock In {\em Conference on Robot Learning}, 2020.

\bibitem[\protect\citeauthoryear{Watters \bgroup \em et al.\egroup
  }{2019}]{watters2019cobra}
Nicholas Watters, Loic Matthey, Matko Bosnjak, Christopher~P Burgess, and
  Alexander Lerchner.
\newblock Cobra: Data-efficient model-based rl through unsupervised object
  discovery and curiosity-driven exploration.
\newblock {\em arXiv preprint arXiv:1905.09275}, 2019.

\bibitem[\protect\citeauthoryear{Xu \bgroup \em et al.\egroup
  }{2019}]{xu2019regression}
Danfei Xu, Roberto Mart{\'\i}n-Mart{\'\i}n, De-An Huang, Yuke Zhu, Silvio
  Savarese, and Li~F Fei-Fei.
\newblock Regression planning networks.
\newblock In {\em Advances in Neural Information Processing Systems}, 2019.

\bibitem[\protect\citeauthoryear{Xu \bgroup \em et al.\egroup
  }{2020}]{xu2020learning}
Tingting Xu, Henghui Zhu, and Ioannis~Ch Paschalidis.
\newblock Learning parametric policies and transition probability models of
  markov decision processes from data.
\newblock {\em European Journal of Control}, 2020.

\bibitem[\protect\citeauthoryear{Ye \bgroup \em et al.\egroup
  }{2020}]{ye2020object}
Yufei Ye, Dhiraj Gandhi, Abhinav Gupta, and Shubham Tulsiani.
\newblock Object-centric forward modeling for model predictive control.
\newblock In {\em Conference on Robot Learning}, 2020.

\bibitem[\protect\citeauthoryear{Zambaldi \bgroup \em et al.\egroup
  }{2018}]{zambaldi2018relational}
Vinicius Zambaldi, David Raposo, Adam Santoro, Victor Bapst, Yujia Li, Igor
  Babuschkin, Karl Tuyls, David Reichert, Timothy Lillicrap, Edward Lockhart,
  et~al.
\newblock Relational deep reinforcement learning.
\newblock {\em arXiv preprint arXiv:1806.01830}, 2018.

\bibitem[\protect\citeauthoryear{Zaragoza \bgroup \em et al.\egroup
  }{2010}]{zaragoza2010relational}
Julio~H Zaragoza, Eduardo~F Morales, et~al.
\newblock Relational reinforcement learning with continuous actions by
  combining behavioural cloning and locally weighted regression.
\newblock {\em Journal of Intelligent Learning Systems and Applications},
  2(02):69, 2010.

\bibitem[\protect\citeauthoryear{Zhu \bgroup \em et al.\egroup
  }{2017}]{zhu2017visual}
Yuke Zhu, Daniel Gordon, Eric Kolve, Dieter Fox, Li~Fei-Fei, Abhinav Gupta,
  Roozbeh Mottaghi, and Ali Farhadi.
\newblock Visual semantic planning using deep successor representations.
\newblock In {\em Proceedings of the IEEE International Conference on Computer
  Vision}, 2017.

\end{thebibliography}
